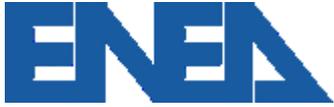

Italian National Agency for New Technologies, Energy and Sustainable Economic Development

http://www.enea.it/en

http://robotica.casaccia.enea.it/index.php?lang=en

# Can underwater robotics technology save submerged Cultural Heritage?

**Ramiro dell'Erba (Corresponding author)**
*ENEA Technical Unit technologies for energy and industry – Robotics Laboratory Via Anguillarese 301, Roma 00123*

**Claudio Moriconi**
*ENEA Technical Unit technologies for energy and industry –Head of the Robotics Laboratory Via Anguillarese 301, Roma 00123*

**Alfredo Trocciola**
*ENEA Technical Unit Technology – Biogeochemistry Laboratory Piazzale E. Fermi, 1, 80055 Portici (NA)*

ABSTRACT

Cultural Heritage can largely profit from the set of technologies that have recently been developed in Submarine Robotics. In this paper we focus on how underwater robotics and related technologies can be used to enhance economical fruition, control, protection and social impact of the cultural heritage. Robots allow on-line experience, in remote locations, realizing the remote museum concept as extension and enhancement of the virtual museum. These solutionspush the cultural tourism beyond actual limits of the sites like the number of simultaneous visitors, the travelling costs,  , the difficulties to access dangerous locations coimng to a true, advanced fruition of the Cultural Heritage goods.

# La tecnologia robotica può salvare I beni culturali sommersi?

ABSTRACT

I beni culturali possono trarre un grande beneficio dalle tecnologie recentemente sviluppate nell'ambito della robotica sottomarina. In quest'articolo focalizzeremo su come la robotica, e le relative tecnologie, possono essere usate per migliorare l'economia, l'esperienza, il controllo, la protezione e l'impatto sociale dei beni culturali. I robot permettono la fruizione online, da località remote, realizzando il concetto di museo remoto, estensione del concetto di museo virtuale. Queste soluzioni, eliminando i costi di viaggio, migliorano e superano gli attuali limiti dei siti, quali il numero di simultaneo di visitatori o la loro posizione in luoghi pericolosi permettendo un salto di qualità nel godimento del bene culturale.

**Introduzione**

La Robotica, con le tecnologie ad essa legate, è stata applicata con successo per diagnosi, ricerca, manutenzione, sorveglianza e fruizione del patrimonio culturale (Fantoni *et al.*, 2013a). Uno dei principali vantaggi offerti è legato alla capacità dei robot di lavorare in ambienti ostili all'uomo, quali i fondali marini, o là dove la presenza umana potrebbe pregiudicare il Bene Culturale. Le stesse tecnologie sono utili strumenti per aumentare la fruizione del bene sia quantitativamente, come aumento della platea, sia qualitativamente per il flusso informativo multimediale e multilivello fornito all'utente, sia esso un esperto o meno. Il concetto di Museo Remoto (European Conference on Research for Protection, Kolar and Strlič, 2010), estensione del concetto di Museo Virtuale, consente una reale presenza dell'osservatore, tramite robot, sul luogo ove si trovi il manufatto; tale concetto di telepresenza deriva direttamente dall' esperienza ENEA sui telemanipolatori per impianti nucleari ove l'operatore, che lavora sugli oggetti del reattore, si trova a distanza di sicurezza ma come se fosse realmente sul posto. A seconda che si operi in ambiente strutturato o meno e in presenza di altre macchine e/o uomini, i robot al lavoro presso il bene culturale devono avere adeguati livelli di autonomia e intelligenza per condividere o meno uno spazio comune, svolgendo compiti cooperando o in antitesi tra i vari attori della scena. In quest'articolo descriveremo l'idea di robotica di sciame sottomarina, una tecnologia che stiamo sviluppando presso il laboratorio di Robotica dell'Enea in collaborazione con team internazionali, ed i vantaggi che può offrire .

**La robotica nei Beni Culturali**

In sintesi la robotica può essere utile nelle seguenti tre tematiche:

- Il concetto di Museo remoto, estensione del museo virtuale.
- Supporto alla restaurazione del bene (Diagnostica e immagini iper realistiche).
- Scoperta di nuovi manufatti.

Molti siti archeologici sono inaccessibili all'uomo o poiché ambienti lontani e/o ostili o per proteggere il bene dalla contaminazione ambientale. Uno o più robot sul posto permettono una visita remota personalizzata, diversa dalla visita virtuale in cui vengono forniti contenuti preformati. Il concetto va oltre la guida turistica del museo sulla quale ci sono state diverse proposte (Burgard *et al.*, 1999) ,(Germak *et al.*, 2015), (Trahanias *et al.*, 2000)

Già alcuni anni fa il concetto è stato introdotto, (European Conference on Research for Protection, Kolar and Strlič, 2010); le tecnologie attuali possono condurre a una visita remota il cui feeling può essere reso imersivo e quindi molto simile a quello di una visita reale anche con costi relativamente limitati e comunque facilmente sostenibili da una struttura museale. Il visitatore del museo remoto, a differenza di quello virtuale, è parte attiva della scena e può interagire per una maggiore fruizione cognitiva e interattiva del bene. Ancora prima della Conferenza del 2010 il concetto è stato sviluppato nel progetto TECSIS (Diagnostic technologies and intelligent systems for the south Italy archaeological sites) (dell'Erba & al., 2006) usando strumentazioni ottiche, analisi ai raggi X, Remote Operated Vehicles (ROV), Underwater Autonomous Vehicles (AUV) e tecniche di realtà aumentata, il progetto comprendeva anche campagne marine effettuate nel Mediterraneo presso siti archeologici sommersi. In particolare si è lavorato presso la città sommersa di Baia (Napoli), la villa di Agrippa Postumo (Sorrento) e un relitto al largo di Punta Imperatore (Ischia). Punti di forza del progetto sono stati l'integrazione di tecnologie provenienti da diversi campi in un corpo omogeneo unitamente allo sviluppo di tecniche, allora totalmente nuove, quali LIBS (Laser induced breakdown spectroscopy) in ambiente sottomarino, l'ablazione laser sottomarina e lo sviluppo di alcune tecniche di controllo dei robot. Il progetto ha prodotto 4 brevetti, 6 campagne di misura,

3 Copyrights di software e più di 100 pubblicazioni. I robot sono in grado di costruire mappe digitali dei siti archeologici e dargli una struttura utilizzando non solo immagini visive ma combinandole con dati di altri strumenti, quali LIF ((laser induced fluorescence), ultraacoustical devices etc. per generare immagini iper realistiche di realtà aumentata che forniscono all'osservatore diversi livelli di lettura contemporaneamente. In particolare, fu eseguita una integrazione di immagini visive con immagini acustiche, prese in contemporanea da strumenti acustici quali Side Scan Sonar (SSS). Inoltre, considerando l'interazione del sistema complessivorobot e pubblico, come nel caso di robot guide, abbiamo anche studiato le abilità sociali dei robot agenti nella vita giornaliera (dell'Erba et al. 2008).

Tecniche di realtà aumentata possono essere di supporto per fornire al restauratore più livelli di visione con analisi (Fantoni *et al.*, 2013b). Unitamente a tecnologie di diagnostica automatica, per il riconoscimento di danni o attacchi ambientali al manufatto, tecniche di identificazione e classificazione intelligente del danno su affreschi, a distanza di 30 metri dal dipinto, sono state utilizzate con successo (Fantoni et al., 2013). Lo scopo era di aiutare il restauratore in un primo riconoscimento dell'esistenza e classificazione del danno, quali ritocchi, restauri precedenti ignoti, e identificazione di componenti chimici (cere, consolidanti, etc..) non precedentemente documentati. Questo è stato effettuato mediante un nuovo tipo di LIF (Laser Induced Fluorescence), tecnica non invasiva capace di distinguere (e poi classificare) la firma spettrale delle sostanze chimiche; un ulteriore vantaggio sta nel fatto di poter operare anche a distanze di 30 metri, evitando di dover essere vicini all'affresco, spesso posto molto in alto.

Il riconoscimento di nuovi manufatti passa attraverso le tecniche di big data processing. L'utilizzo, contemporaneo, di più sensori in uno stesso volume e la loro correlazione richiede calcoli intensi unitamente all'utilizzo di tecniche di Deep Learning e comprensione del contesto (Conte *et al.*, 2009), (Conte *et al.*, 2010). Questo permette la realizzazione di mappe dinamiche multilivello sulle quali si può inserire un lavoro di contestualizzazione semantica per la classificazione intelligente. Il riconoscimento degli oggetti è quindi facilitato utilizzando isomorfismi tra sub mappe di diverso livello gerarchico. Un'altra applicazione di tecnologia robotica utile è il "ritorno di forza sulla mano" della persona che controlla il robot, tecnica indispensabile per i robot che operano, sotto il diretto controllo dell'uomo, negli impianti nucleari e che nel caso della fruizione delle opere d'arte, unitamente a telecamere tridimensionali, permette all'uomo una migliore percezione, anche tattile del manufatto, (Scaradozzi *et al.*, 2014).

I robot impiegabili in ambiente subacqueo sono di diverso tipo dai ROV, AUV, Glider etc…ognuno di essi presenta dei vantaggi e degli svantaggi. Ad esempio, un ROV presenta il vantaggio di poter trasmettere una grande quantità di dati, attraverso il cavo di collegamento, e un'interazione in tempo reale con l'operatore. D'altra parte, lo stesso cavo richiede una nave appoggio con costi e logistiche superiori a quelle di un AUV, che ha nelle comunicazioni il suo principale punto debole. Interessanti sono gli ibridi, che mediante l'uso di uno scafo robotizzato che si muove sulla superficie marina, mantengono il contatto (tipicamente mediante modem acustico) con un AUV in immersione e via radio con l'operatore umano a terra.

Tra le diverse opzioni possibili il nostro laboratorio ha scelto di considerare strategica <u>la robotica di sciame</u>

**La Robotica di Sciame e le sue motivazioni**

Un nuovo modo di proteggere e fruire dei Beni Culturali siti in ambienti sottomarini, di cui il nostro Paese è ricchissimo, riguarda la Robotica di Sciame. In essa un gruppo organizzato (dell'Erba et al. 2008) di AUV economici, di piccole dimensioni (MAUV) ed equipaggiati con una sensoristica minima e videocamere sono in grado di esplorare grandi aree marine simultaneamente, aumentando la probabilità di scoperta. Gli sciami però intervengono anche nel monitoraggio continuo del Bene, nella sua protezione contro le modifiche ambientali, nella sua difesa contro il saccheggio operato da organizzazioni criminali e, come detto, nella sua fruizione.

I vantaggi di lavorare con uno sciame, rispetto al singolo robot sono i seguenti [5]:

1) Parallelismo delle operazioni, quindi maggiore velocità;
2) Robustezza del sistema e salvaguardia della missione in caso di perdita di uno o più elementi;
3) Possibilità di avere più "punti di vista" (nella sua più larga accezione, non solo visiva) contemporaneamente;
4) Possibilità di concentrare l'esplorazione dove serve (Ad esempio in caso di scoperta di un manufatto);
5) Possibilità di realizzare un network sottomarino di comunicazione.
6) possibilità per l'uomo di controllare un sistema complesso interfacciandosi con un'unica creatura: L'operatore assegna il compito ed poi è lo sciame a effettuare il "job sharing" tra i vari elementi;

Gli svantaggi invece sono:

7) Una relativa maggiore sofisticazione del sistema di controllo che impone ancora un certo livello di ricerca e sviluppo
8) Necessità di realizzare protocolli di comunicazione avanzati affinché sia garantita una comunicazione in grado di assicurare l'integrità dello sciame.

La semplicità del singolo veicolo, indispensabile per contenere i costi, è largamente recuperata grazie alla molteplicità dei sistemi sensoriali che possono integrarsi tra loro per fornire misure ambientali più precise e dettagliate. L'operatore umano comunica con lo sciame come se fosse una creatura unica e poi sarà il sistema di controllo distribuito che controlla la configurazione e la navigazione a regolare il comportamento dei singoli elementi dello sciame. Non esistono, al momento in cui si scrive, soluzioni commerciali di questo tipo. L'impiego di sciami di robot è la frontiera verso la quale si stanno indirizzando alcuni dei principali laboratori che lavorano nella robotica sottomarina.

Già nel 2008, il Laboratorio di Robotica dell'Enea (IDRA) ha lanciato l'innovativo concetto delle "self-organising complex creatures" nell'ambito del progetto Harness, finanziato dall'IIT. Harness si basava sul paradigma di sciame e ha provato a superare uno dei problemi più gravi di un AUV che lavori da solo: la limitata capacità di comunicazione con un supervisore umano, in ambito sottomarino. IDRA ha sviluppato un modello di AUV molto economico (15 Keuro come prezzo base per il prototipo), ottimizzando ogni dettaglio, la cui caratteristica principale è di poter lavorare in sciame; lo sciame ha un innovativo sistema di controllo per gestire sia la sua configurazione geometrica che il flusso dati del canale di comunicazione opto acustico. Tale lavoro ha prodotto anche un brevetto legato alla capacità del sistema di robot di avere una costante consapevolezza della sua dislocazione rispetto all'ambiente.Si realizza in pratica una vera e propria internet sottomarina. Un'altra caratteristica del mezzo è di avere un solo motore e diverse coppie di pinne, un'anteriore e due posteriori, per attuare i cambiamenti di quota e direzione. La presenza di due coppie permette al robot manovre particolarmente agili e perfino qualcosa di molto simile all'hovering, ovvero di rimanere stazionario su una zona. Questo avviene mettendo in contrasto la spinta delle pinne e provocando uno stallo del veicolo, con il vantaggio di averlo anche inclinato rispetto al fondo per una migliore visuale. Questa manovra di volo in altri robot è realizzata, tipicamente, mediante l'uso di almeno altri quattro motori ausiliari con un consistente incremento dei costi.

L'operatore può essere a terra se uno degli AUV sta in superficie e funge da anello di collegamento tra il network sottomarino e la base. Questo riduce drasticamente i costi logistici, facendo a meno della nave appoggio.

Il punto più forte a favore dello sciame è di certo la creazione di un network sottomarino che superi i problemi di comunicazione. Si tratta essenzialmente di una rete multi-hop a geometria variabile. L'operatore "visualizza" (le virgolette stanno poiché il dato non è necessariamente visivo, ma dipende dal sensore) le scene provenienti da differenti MAUV. La configurazione dello sciame si adatta a seconda delle esigenze per cui se è necessario esplorare un grande volume di mare acquisirà una configurazione ellissoidale, se una superficie gli elementi si spanderanno su un quadrato mentre se è necessario comunicare con una banda larga assumeranno una configurazione a tubo. In caso di area sospetta la concentrazione degli elementi aumenterà. E' molto importante porre l'accento su come la filosofia dello sciame sia molto diversa da quella del singolo AUV; è la cooperazione che permette una raccolta dati qualitativamente e quantitativamente superiore a quella del singolo AUV pur usando una sensoristica meno sofisticata sul singolo elemento, rispetto ai singoli AUV di più grande complessità e sofisticazione, in grado di operare da soli.

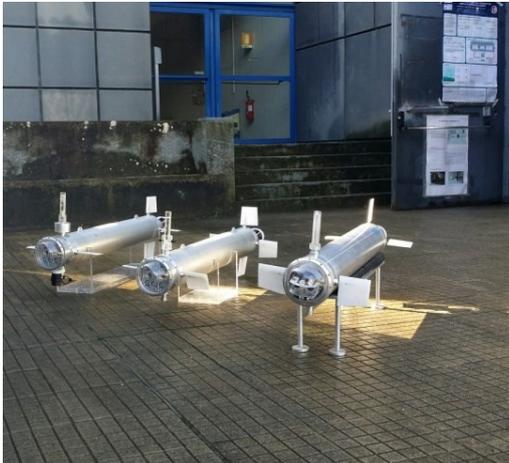
**Figura 2 - Prototipi di MAUV**

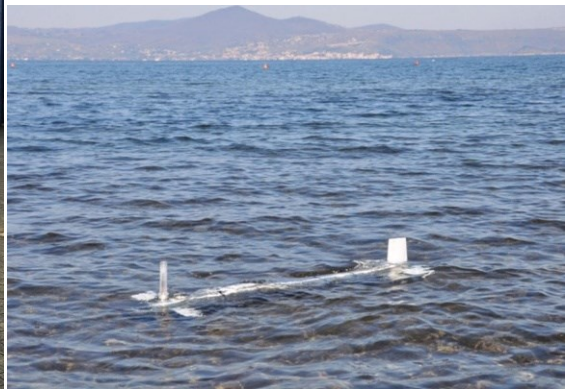
**Figura 1 – Prototipo in sperimentazione nel lago di Bracciano**

Il laboratorio IDRA sta lavorando da diversi anni su questo. L'operatore umano vede lo sciame come se fosse un unico oggetto e come tale si interfaccia con esso superando la difficoltà di comunicare ordini a diversi elementi. La percezione distribuita dello sciame, unitamente a tecniche di comunicazione e calcolo dedicate, permette una maggiore comprensione del contesto ambientale. Va sottolineato che anche gli operatori subacquei professionali hanno difficoltà a comprendere la natura di certi oggetti se prima non li vedono da diversi punti di vista. In Figura 2 e Figura 1 sono mostrati alcuni prototipi, realizzati dal laboratorio ed in via di sperimentazione presso il lago di Bracciano. Le loro caratteristiche sono le seguenti: profondità massima 100 m, velocità massima 2 nodi, peso circa 20Kg, dimensioni 1,30x0,25 m. L'Equipaggiamento standard prevede una camera stereoscopica, sonar, accelerometri, bussola e profondimetro.

Lo sciame si avvale, per le comunicazioni subacquee, di un sistema ibrido ottico-acustico (Figura 3) realizzato e brevettato in ENEA presso il Laboratorio IDRA. La comunicazione sottomarina è oggi largamente affidata a sistemi modem di tipo acustico, che usano boe o navi appoggio di notevoli dimensioni, e basse velocità di comunicazione ed è molto influenzata dalle caratteristiche dell'area di operazione. La trasmissione di foto e l'interazione con l'operatore umano sono spesso impossibili;

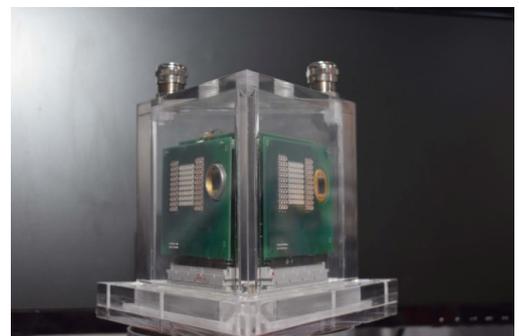
**Figura 3 - Prototipo di modem ottico realizzato dal Laboratorio di Robotica dell' ENEA**

l'impiego di sciami di robot dotati di sistemi di trasmissione ottica rende queste applicazioni fattibili, se le condizioni ambientali lo permettono. Il Laboratorio ENEA lavora in collaborazione con il Dipartimento di Ingegneria Elettronica dell'Università di Tor Vergata per la parte specificamente legata ai protocolli di comunicazione; occorre infatti un protocollo dedicato che utilizzi i due canali, ottico e acustico, alternativamente o in cooperazione e che determini la configurazione geometrica dello sciame ottimizzando la trasmissione compatibilmente con le richieste di esplorazione. Lo scambio dati tra gli elementi dello sciame, e la successiva elaborazione a bordo, è indispensabile affinché si possa considerarlo un unico oggetto.

**L'esperienza di Sinuessa e sua possibile robotizzazione**

Un esempio concreto di come la tecnologia robotica può contribuire a salvare i beni culturali sommersi è fornita dall'esperienza ENEA acquisita in Campania sul sito archeologico di *Sinuessa.* L'Enea a partire dal 2012 è stata coinvolta dai comuni di Sessa Aurunca e, in seguito, di Mondragone per una valorizzazione dell'area costiera mediante il turismo sostenibile. Per tale motivo sono state effettuate delle ricerche marine nel golfo di Gaeta dove era probabile l'esistenza di un approdo sommerso della colonia romana di *Sinuessa* risalente al III secolo a.C. (Pennetta et al., 2016,201, Trocciola et. al., 2013) ed esattamente dove erano i resti di strutture portuali (*pilae*): ad una distanza di circa 650 m dalla costa e ad una profondità di -8 m, all'interno di una depressione del fondale marino. Le acque marine di Sinuessa sono molto torbide per la sospensione dei sedimenti ed il consistente apporto dei nutrienti dai corsi d'acqua fluviali del Garigliano e Volturno, che hanno scoraggiato l'esplorazione subacquea, ma anche preservato il sito dalla diffusa attività clandestina.

In considerazione di ciò, già nel 2013 l'ENEA ha effettuato in questo sito una prima applicazione di tecnologie robotiche con l'impiego del *Side Scan Sonar*. Il Sonar a scansione laterale impiegato, mod. 3900 della L3 KLEIN Associates, con una frequenza acustica ad alta risoluzione (450 kHz), è stato pilotato con tecniche innovative per un rilievo subacqueo sotto costa a basse profondità. L'indagine elettroacustica ha consentito di acquisire la mappatura dei fondali marini (mosaicatura dei sonogrammi, fig.4) in modo da ricostruire un assetto geomorfologico di un tratto costiero significativo di forma rettangolare di dimensioni pari a 1,2 x 1,5 km del territorio comunale di Sessa Aurunca e di interpretare le strutture geomorfologiche e geoarcheologiche evidenziate dal Side Scan Sonar durante il rilievo geofisico con dei rilievi di dettaglio in immersione per la verifica dei target (fig.5, Trocciola et. al.,2017) .

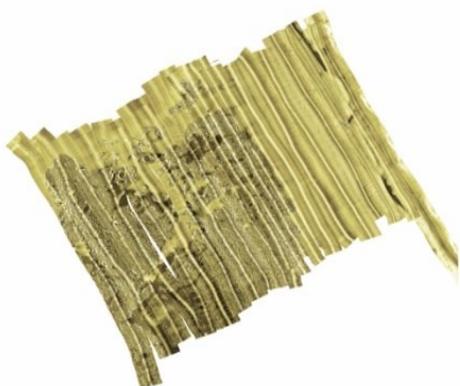
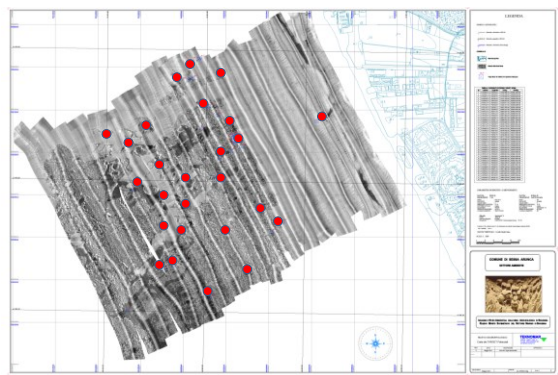

**Figura 4 – Mosaicatura dei sonogrammi con il Side Scan Sonar (immagine ripresa dalla copertina del IVnumero). del 2013 di Archeomatica**

**Figura 5 – Mosaicatura dei sonogrammi con indicazione dei target ispezionati con immersioni dirette**

Il rilievo con il *Side scan Sonar* è stato scelto per l'area archeologica sommersa di *Sinuessa* perché fornisce una accurata indagine non invasiva (preserva l'integrità dei resti archeologici), si effettua in tempi rapidi ed è relativamente economica. Le esplorazioni geofisiche eseguite con questa tecnologia riducono i costi di indagine soprattutto quando si esplorano ambienti ostili e/o inaccessibili, in questo caso vengono limitate le ulteriori perlustrazioni dirette previste per il controllo da parte degli operatori subacquei.

La scelta di elaborare un itinerario subacqueo (Trocciola, 2017), in linea con la Convenzione UNESCO del 2001 per la protezione del patrimonio culturale subacqueo, è stata fatta per una maggiore fruizione dei beni culturali subacquei e la loro preservazione in situ, evitando con l'installazione di sistemi di controllo e di monitoraggio una migliore fruizione e l'uso indiscriminato e distruttivo arrecato dal prelievo di reperti dai fondali. Il braccio di mare antistante il litorale di Sessa Aurunca e Mondragone, potrà divenire con questa soluzione ipotizzata un grande **museo diffuso sommerso**, dove le testimonianze dell'uomo del passato convivono senza alterare il contesto originario della giacitura geologica insieme ai reperti archeologici

I reperti ne sono la testimonianza. Inoltre, tale iniziativa incentiva l'offerta culturale delle amministrazione locali e stimola dei comportamenti corretti ed azioni consapevoli per una valorizzazione partecipativa e sostenibile del turismo.

Il prossimo passo da parte dell'ENEA nell'approccio scientifico del patrimonio culturale sommerso di Sinuessa è di applicare la **Robotica di Sciame** sia nella fase di indagine sui fondali di *Sinuessa* che nella fase di tutela. In particolare, la rapida evoluzione tecnologica nelle prospezioni sottomarine con la moltiplicazione di sonar a basso costo aumenta la probabilità di individuare altri ritrovamenti archeologici in mare ed in tempi estremamente ridotti

Inoltre, il patrimonio archeologico subacqueo di *Sinuessa* si può tutelare nel tempo con gli stessi **sciami di robot** utilizzati nei musei remoti. La musealizzazione remota, con la visione tridimensionale della realtà virtuale impiegando dei robot telecomandati, sul fondale marino di Sinuessa permettono l'osservazione delle sue peculiarità archeologiche e naturalistiche. Così adoperando gli stessi robot sarà possibile effettuare a costi contenuti anche un efficace sistema di monitoraggio remoto e di controllo di continuo per la salvaguardia del patrimonio culturale subacqueo.

**Conclusioni e lavoro futuro**

Il nostro Paese dispone del più grande museo aperto nel Mediterraneo lungo le coste italiane sui beni archeologici sommersi e, se utilizzeremo al meglio, le ultime tecnologie robotiche di **sciame sviluppate ed in via di perfezionamento all'ENEA per** i beni culturali, potremo contribuire alla sua valorizzazione, fruizione e tutela. La robotica di sciame, insieme alle tecnologie ad essa correlate, può costituire un punto di svolta per un rilancio e un nuovo tipo di fruizione dei Beni Culturali situati sui fondali marini. Le tecnologie e i costi ad esse relativi sono ormai maturi per applicazioni in siti marini già documentati e possono costituire anche un motivo di rilancio per le economie locali, che hanno già mostrato interesse per questo tipo di attività.

**Bibliografia**